%% file: sample-sigconf.tex
\documentclass[sigconf,authorversion]{acmart}
\usepackage{enumitem}
\usepackage{hyperref}
\usepackage{multirow}
\usepackage{amsthm}
\usepackage{array}
\usepackage{subfigure}
\usepackage{graphicx}
\setlist[enumerate]{align=left}
\usepackage{algorithm}
\usepackage{algorithmic}
\usepackage{subcaption}
\usepackage{xcolor}

\usepackage[multiple]{footmisc}

\AtBeginDocument{%
  }

\copyrightyear{2026}
\acmYear{2026}
\setcopyright{acmlicensed}\acmConference[KDD '26]{Proceedings of the 32nd SIGKDD Conference on Knowledge Discovery and Data Mining}{August 9--13, 2026}{Jeju, Korea}
\acmBooktitle{Proceedings of the 32nd SIGKDD Conference on Knowledge Discovery and Data Mining (KDD '26), August 9--13, 2026, Jeju, Korea}
\acmDOI{XXXXXXX.XXXXXXX}
\acmISBN{978-1-4503-XXXX-X/2018/06}

\begin{document}

\title{Intelligent Approval of Access Control Flow in Office Automation Systems via Relational Modeling}

\author{Dugang Liu}
\affiliation{%
  \institution{College of Computer Science and Software Engineering, Shenzhen University}
  \city{Shenzhen}
  \country{China}}  
\email{dugang.ldg@gmail.com}

\author{Zulong Chen}
\affiliation{%
\institution{Alibaba Group Holding Limited}
\city{Hangzhou}
\country{China}}
\email{zulong.czl@alibaba-inc.com}

%%corresponding author
\author{Chuanfei Xu}
\authornote{*Corresponding authors}
\affiliation{%
\institution{Guangdong Laboratory of Artificial Intelligence and Digital Economy (SZ)}
\city{Shenzhen}
\country{China}}
\email{xuchuanfei@gml.ac.cn}

%%corresponding author
\author{Jiaxuan He}
\authornotemark[1]
\affiliation{%
\institution{Alibaba Group Holding Limited}
\city{Hangzhou}
\country{China}
}
\email{hejiaxuan.hjx@alibaba-inc.com}

\author{Yunlu Ma}
\affiliation{
\institution{Alibaba Group Holding Limited}
\city{Hangzhou}
\country{China}
}
\email{yunlu.myl@alibaba-inc.com}

\author{Jia Xu}
\affiliation{
\institution{Cyberspace Institute of Advanced Technology, Guangzhou University}
\city{Hangzhou}
\country{china}
}
\email{xujia@gzhu.edu.cn}

\author{Xing Tang}
\affiliation{%
  \institution{Shenzhen Technology University}
  \city{Shenzhen}
  \country{China}
}
\email{xing.tang@hotmail.com}

\author{Xiuqiang He}
\affiliation{%
  \institution{Shenzhen Technology University}
  \city{Shenzhen}
  \country{China}
}
\email{hexiuqiang@sztu.edu.cn}

\renewcommand{\shortauthors}{Dugang Liu, et al.}

\input{sections/abstract}




\maketitle

\input{sections/introduction}
\input{sections/formulation}

\input{sections/method}
\input{sections/experiments}
\input{sections/relatedwork}
\input{sections/conclusion}

\bibliographystyle{ACM-Reference-Format}
\bibliography{sample-base}

\end{document}

%% file: sections/abstract.tex
\begin{abstract}
Office automation (OA) systems play a crucial role in enterprise operations and management, with access control flow approval (ACFA) being a key component that manages the accessibility of various resources. However, traditional ACFA requires approval from the person in charge at each step, which consumes a significant amount of manpower and time. Its intelligence is a crucial issue that needs to be addressed urgently by all companies. In this paper, we propose a novel relational modeling-driven intelligent approval (RMIA) framework to automate ACFA. Specifically, our RMIA consists of two core modules: (1) The binary relation modeling module aims to characterize the coupling relation between applicants and approvers, and provide reliable basic information for ACFA decision-making from a coarse-grained perspective. 2) The ternary relation modeling module utilizes specific resource information as its core, characterizing the complex relations between applicants, resources, and approvers, and thus provides fine-grained gain information for informed decision-making. Then, our RMIA effectively fuses these two kinds of information to form the final decision. Finally, extensive experiments are conducted on two product datasets and an online A/B test to verify the effectiveness of RMIA.
\end{abstract}

%% file: sections/introduction.tex
\section{Introduction}\label{sec:introduction}
\begin{figure}[htbp]
\centering
\includegraphics[width=1.\linewidth]{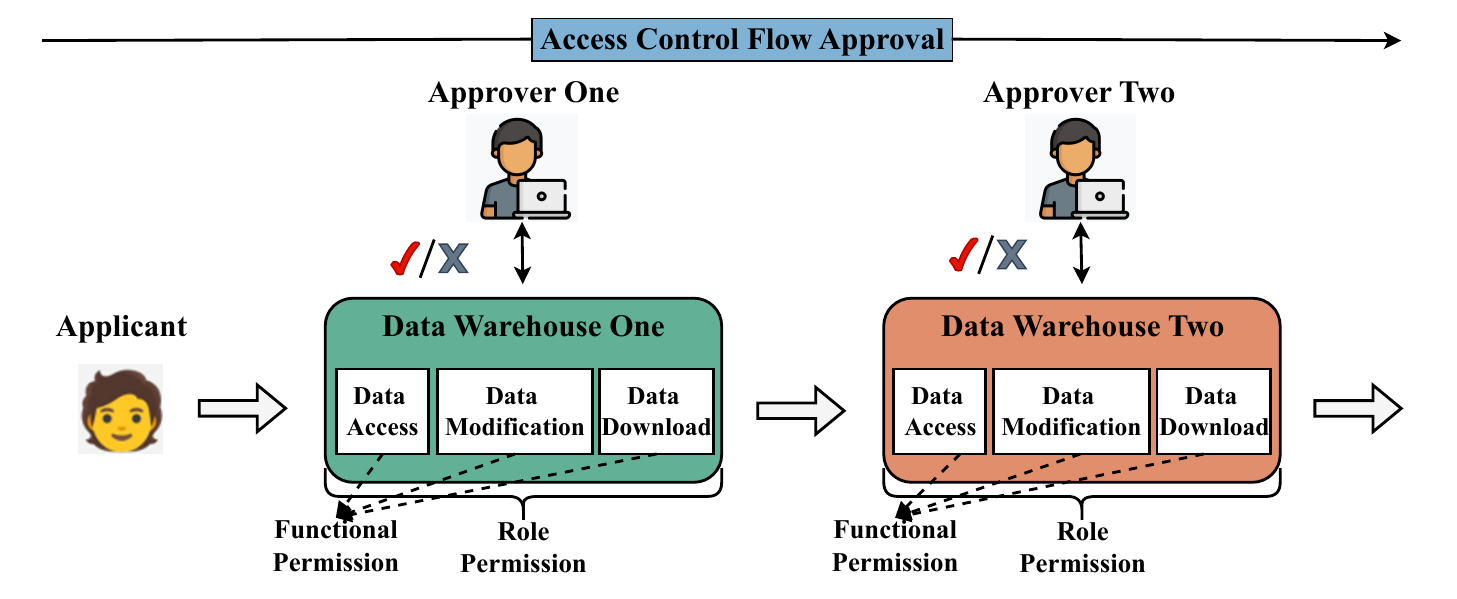}
\caption{An example of an access control flow approval (ACFA) process, where each data resource application is considered a step. The applicant initiates a permission application for a series of data resources due to specific business needs, and each step is reviewed and approved by the corresponding approver. Note that there are different permission types depending on the functional scope.}
\label{fig:1}
\end{figure}

% ==
Office automation (OA) systems have become integral to the operation and management of modern enterprises, significantly enhancing the efficiency of administrative, financial, and business operational processes through paperless and easily traceable services~\cite{olson1982impact,bhuyar2016design,van2012process,coombs2020strategic,shlomov2024towards,kedziora2023turning}.
% ==
Access control flow approval (ACFA) is one of the most frequently used components in OA, which aims to effectively manage the accessibility of different employees to specific resources, thereby enhancing the security of those resources.
% ==
As shown in Figure~\ref{fig:1}, in each step of ACFA, the applicant initiates an application for permission to a specific resource. Then, each responsible approver verifies and decides on the application.
% ==
Note that permissions are of different types, such as functional permissions that only activate certain sub-functions and role permissions that activate all functions~\cite{kumar2002context}.

% =======================================================================================
%=======================================================================================

% ==
However, the current ACFA, which relies on manual decision-making, will impose a non-negligible processing burden on approvers as the volume of permission applications increases, a common occurrence in large enterprises.
% ==
Additionally, this may render the processing time of each step in ACFA uncontrollable, resulting in a significant time cost. For example, based on our business scenarios from the past 7 months, the average time for each approval task is approximately 10 hours.
% ==
Given these issues, we are motivated to explore the effective integration of artificial intelligence (AI) with ACFA to achieve accurate automated intelligent approval~\cite{paoki2024artificial,pennathur2024future}.
% ==
In fact, a large portion of permission applications in the historical approval data of most companies are typically low-risk in nature, resulting in a high approval rate, such as the renewal permission application. This also provides a favorable environment for the feasibility of ACFA intelligent approval.

% ==
To the best of our knowledge, unlike other intelligent approval tasks~\cite{lee2024submodular,tang2024enabling,purificato2023use,long2024sphere}, intelligent approval for ACFA has received less attention in the existing literature.
% ==
To this end, this paper proposes a novel relational modeling-driven intelligent approval (RMIA) framework for automated ACFA to fill this gap.
% ==
Our RMIA consists of two core modules: the binary relation modeling module and the ternary relation modeling module, featuring a hierarchical decision-making mechanism to facilitate more informed approval decisions.
% ==
Specifically, the former utilizes the coupling relation between applicants and approvers to provide a reasonable basis for decision-making; the latter introduces resources as hubs and leverages the complex relation between applicants, resources, and approvers to provide fine-grained gain information for decision-making.
% ==
These two pieces of information are then aggregated to form the final approval decision.
% ==
Finally, comparative experiments on two product datasets and an online A/B test verify the effectiveness of RMIA.

%% file: sections/formulation.tex
\section{Problem Formulation}\label{sec:problem}
% ==
In this section, we first formulate the definition and necessary notation of the intelligent approval task of ACFA.
% ==
We regard each step in ACFA as the smallest atom to construct training instances, i.e., each training instance includes a specific applicant's permission application for a particular resource and the corresponding approval result made by an approver.
% ==
Therefore, the training instance in the intelligent approval task can be denoted as $\mathcal{D}= \{(\mathbf{x}_i,\mathbf{r}_i,\mathbf{\overline{x}}_i,\mathbf{t}_i,\mathbf{s}_i,\mathbf{p}_i,y_i) \}_{i=1}^n$, where for the $i$-th instance, $\mathbf{x}_i$ is the applicant feature vector, $\mathbf{r}_i$ is the resource feature vector, $\mathbf{\overline{x}}_i$ is the approver feature vector, $\mathbf{t}_i$ is the application-related text feature vector, $\mathbf{s}_i$ is the historical application statistics vector, $\mathbf{p}_i$ is the applicant-approver affinity feature vector, and $y_i\in \{0,1\}$ is the corresponding decision label, i.e., pass or fail.
% ==
Based on the above training instance set, the intelligent approval task aims to train a model $\hat{y}_i=f(\mathbf{x}_i,\mathbf{r}_i,\mathbf{\overline{x}}_i,\mathbf{t}_i,\mathbf{s}_i,\mathbf{p}_i\mid \theta)$ to make decisions for each permission application in ACFA, where $f(\cdot)$ is the feature-to-label mapping function implemented by different models, and $\theta$ is model parameters.
% ==
% ==
In practice, we use the cross-entropy function to optimize this intelligent approval model,
\begin{equation}\label{eq:1}
\mathcal{L}=\sum_{i=1}^{n}l(y_i, \hat{y}_i),
\end{equation}
where $l(\cdot)$ and $\hat{y}_i$ are the cross-entropy loss and the predicted label of the intelligent approval model, respectively.

%% file: sections/method.tex
\section{The Proposed Framework}\label{sec:method}
% ==
In this section, we will first describe the overall framework of our RMIA.
% ==
Then, we will follow the training pipeline and provide a detailed introduction to each module in RMIA, including the core binary and ternary relation modeling.

\subsection{Framework Overview}
\begin{figure*}[htbp]
\centering
\includegraphics[width=1.\linewidth]{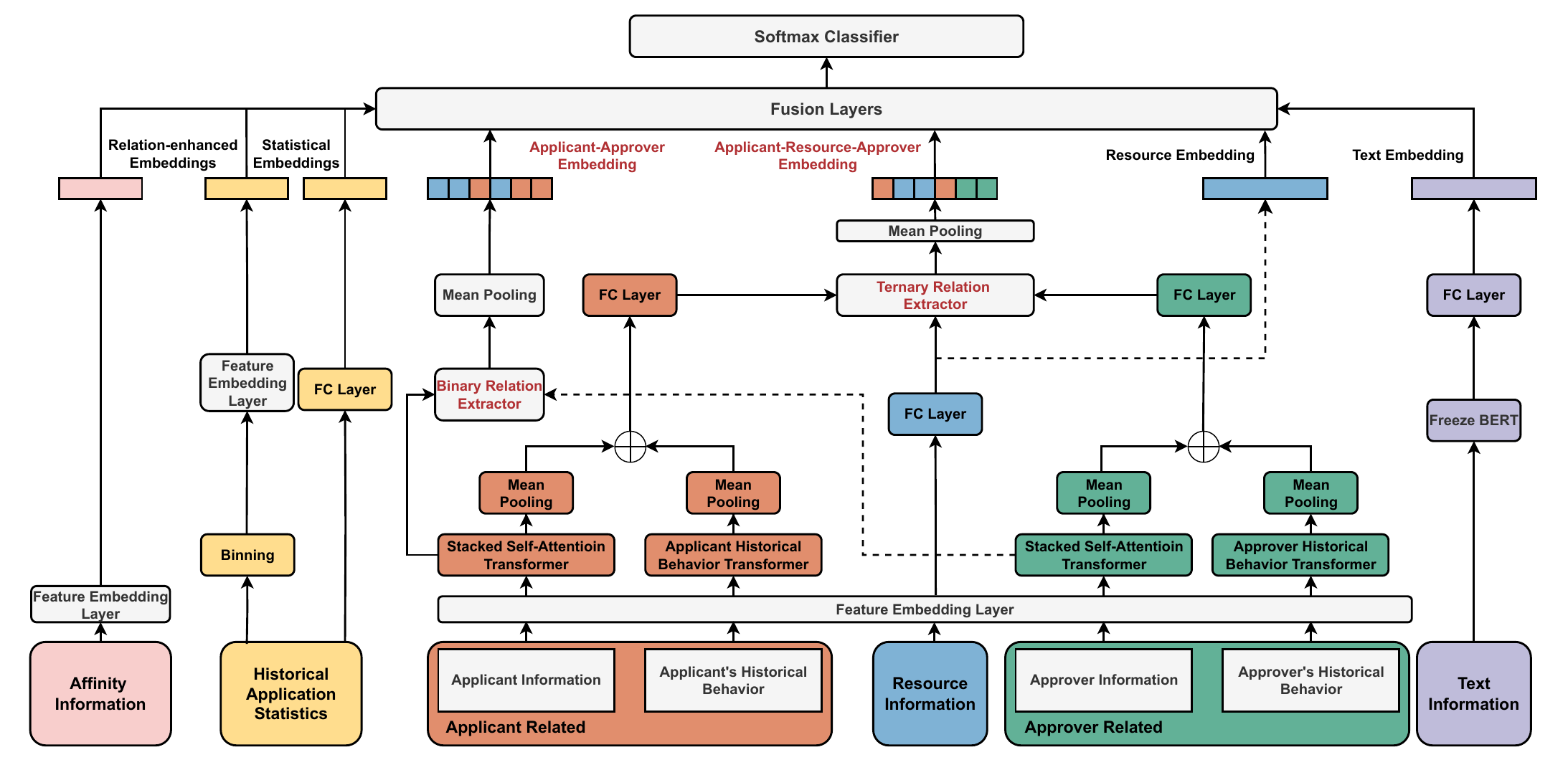}
\caption{The architecture of our Relational Modeling-driven Intelligent Approval (RMIA) framework, which primarily includes a binary relation extractor for applicants and approvers, and a ternary relation extractor for applicants, resources, and approvers.}
\label{fig:2}
\end{figure*}

% ==
As shown in Figure~\ref{fig:2}, our RMIA takes three primary sources as input: applicant features, resource features, and approver features.
% ==
The features of applicants and approvers include their identity information, such as work number, position, and department, and historical behavior, such as the recent $k$ application records and approval records.
% ==
These two pieces of information will pass through the transformer to obtain identity embedding and behavior embedding, respectively, and they will be fused to obtain applicant and approver embeddings (marked in orange and green).

% ==
Then, RMIA’s binary relation modeling module will utilize a customized extractor to mine the connections between applicants and approvers based on their identity embeddings and obtain applicant-approver embeddings that characterize the binary relation, which is a fundamental information source for subsequent decision-making.
% ==
To further enhance the accuracy of binary relation modeling, we explicitly introduce affinity information (marked in pink) between applicants and approvers, such as their organizational distance and historical interactions (whether they have attended meetings together or participated in the same business), and transform it into a relation-enhanced embedding.

% ==
Next, after passing resource features through a fully connected layer to obtain resource embeddings (marked in blue), RMIA's ternary relation modeling module introduces a customized extractor based on the embeddings of applicants, resources, and approvers to comprehensively capture the complex coupling relations between them and obtain the applicant-resource-approver embedding corresponding to the ternary relation.
% ==
This will serve as fine-grained gain information for subsequent decision making.
% ==
Similarly, to improve the modeling of ternary relations, we explicitly introduce historical application statistics (such as the number of applications and approval rates for the same applicant-resource pair; the number of applications and approval rates for the same applicant-resource-approver pair, etc.) to obtain a set of statistical embeddings for enhanced ternary relations.

% ==
Finally, the embeddings of the above parts will be merged into a fusion layer and then passed through a prediction layer to obtain the predicted decision label.
% ==
This means that RMIA will effectively weigh the basic information and gain information corresponding to binary and ternary relations to make reasonable hierarchical decision-making thinking.
% ==
Note that since permission applications may contain some optional text information, such as application reasons, permission description, and approval summary, RMIA directly obtains text embeddings based on the frozen BERT model and fuses them with other embeddings.

% =======================================================================================
%=======================================================================================

\subsection{Training}
% ==
In this subsection, we describe each module in detail based on the training process.

\subsubsection{The Feature Encoder Module}\label{subsubsec: feature}
% ==
Given a current approval instance $\mathbf{z}=(\mathbf{x},\mathbf{r},\mathbf{\overline{x}},\mathbf{t},\mathbf{s},\mathbf{p},y) $, to better represent and utilize its features, we first need to encode different types of features into a low-dimensional and dense real-valued vector through a feature embedding layer with an embedding table $\mathbf{E}\in\theta$. 

% ==
For the applicant feature vector $\mathbf{x}$, it can be divided into two parts: the identity information subset $\mathbf{x}^{id}$ with $d_1$ features and the historical behavior subset $\mathbf{x}^{hb}$ with recent $k$ application records, i.e., $\mathbf{x}=\left[\mathbf{x}^{id}_{1},\cdots,\mathbf{x}^{id}_{d_1},\mathbf{x}^{hb}_{1},\cdots,\mathbf{x}^{hb}_{k}\right]$.
% ==
For the identity information subset, we use the \textit{lookup} operation to obtain their feature embeddings from the initialized embedding table $\mathbf{E}$, denoted as $\mathbf{e}^{\mathbf{x}^{id}}=\left[\mathbf{e}^{\mathbf{x}^{id}}_{1},\cdots,\mathbf{e}^{\mathbf{x}^{id}}_{d_1}\right]$.
% ==
For each historical behavior, it contains a complete application record for the current applicant, i.e., the corresponding resource, approver, and decision result $\left(\mathbf{r}^{hb},\mathbf{\overline{x}}^{hb},y^{hb}\right)$.
% ==
Therefore, we we also first obtain the corresponding embeddings for each feature in the record from the embedding table $\mathbf{E}$ and concatenate them to get the embedding for each historical behavior, i.e., $\mathbf{e}^{\mathbf{x}^{hd}}_{j}=concat\left(\mathbf{e}^{\mathbf{r}^{hb}},\mathbf{e}^{\mathbf{\overline{x}}^{hb}},\mathbf{e}^{y^{hb}}\right)$ and $\mathbf{e}^{\mathbf{x}^{hd}}=\left[\mathbf{e}^{\mathbf{x}^{hd}}_{1},\cdots,\mathbf{e}^{\mathbf{x}^{hd}}_{k}\right]$.

% ==
Similarly, for the approver feature vector $\mathbf{\overline{x}}$, we can follow the same process to obtain the corresponding identity information and historical form embedding, i.e., $\mathbf{e}^{\mathbf{\overline{x}}^{id}}=\left[\mathbf{e}^{\mathbf{\overline{x}}^{id}}_{1},\cdots,\mathbf{e}^{\mathbf{\overline{x}}^{id}}_{d_2}\right]$ and $\mathbf{e}^{\mathbf{\overline{x}}^{hd}}=\left[\mathbf{e}^{\mathbf{\overline{x}}^{hd}}_{1},\cdots,\mathbf{e}^{\mathbf{\overline{x}}^{hd}}_{k}\right]$.
% ==
Note that since the number of identity information features of the approver may be different from that of the applicant, we use $d_2$ to represent the number of approver features. 
% ==
In addition, for the approver, historical behavior will become an approval record, i.e., the corresponding applicant, resource, and decision result $\left(\mathbf{r}^{hb},\mathbf{x}^{hb},y^{hb}\right)$.

% ==
For the resource feature vector $\mathbf{r}$, we obtain embeddings for features, such as resource ID and permission type, from the embedding table $\mathbf{E}$ separately, and concatenate them to obtain the resource feature embedding $\mathbf{e^{\mathbf{r}}}$.

% ==
For the application-related text feature vectors $\mathbf{t}$, including the application reason $\mathbf{t}_r$, permission description $\mathbf{t}_d$, and approval summary $\mathbf{t}_s$, we use a frozen BERT~\cite{devlin2019bert} to encode each type of text.
% ==
Then, we concatenate the output CLS features and use a fully connected layer to map them to obtain the final text embedding $\mathbf{e^t}$.
\begin{equation}\label{equ:2}
\mathbf{T}=\mathrm{BERT}\left(\mathbf{t}_r\right)_{[CLS]}\oplus\mathrm{BERT}\left(\mathbf{t}_d\right)_{[CLS]}\oplus\mathrm{BERT}\left(\mathbf{t}_s\right)_{[CLS]},
\end{equation}
\begin{equation}\label{equ:3}
\mathbf{e^t}=\mathbf{W}_{1}\mathbf{T}+\mathbf{b}_{1},
\end{equation}
where $\mathbf{W}_{1}$ and $\mathbf{b}_{1}$ are trainable parameters.

% ==
For the affinity information feature vector $\mathbf{p}$ used to enhance binary relation modeling, we use categorical features (e.g., five categories) to identify the organizational distance and historical interaction (such as whether they attended meetings together or participated in the same business) between applicants and approvers.
% ==
We then convert this feature vector into a relation-enhanced embedding $\mathbf{e}^{\mathbf{p}}$ directly from the embedding table $\mathbf{E}$.

% ==
For the historical application statistical information feature vector $\mathbf{s}$ of enhanced ternary relation modeling, it mainly incorporates the historical association attributes between resources and applicants or approvers, such as the historical number of applications and approval rates for the same applicant-resource pair; the historical number of applications and approval rates for the same applicant-resource-approver pair, etc.
% ==
For the features of the number of applications, we first convert them into categorical features using a binning operation, and then obtain the statistical embeddings of these types of features from the embedding table.
% ==
We concatenate all embeddings of this type in the historical application statistics and denote it as $\mathbf{e}^{\mathbf{s}_{n}}$.
% ==
For the approval rate feature, we first perform a log normalization operation on it and use a fully connected layer to obtain this type of statistical embedding. 
% ==
We concatenate the embeddings of all approval rate types in the historical application statistics and denote it as $\mathbf{e}^{\mathbf{s}_{r}}$.

\subsubsection{The Binary Relation Modeling Module}\label{subsubsec: binary}
% ==
In ACFA's approval process, an underlying factor is the affinity between the applicant and the approver.
% ==
On the one hand, an approver who belongs to the same department as the applicant or has frequent historical interactions with the applicant may be more familiar with the applicant and may have a higher probability of approving his/her application.
% ==
On the other hand, in the above situation, the resources involved in the permission application are more likely to be common resources within the department or a specific business, thus having a higher approval rate.
% ==
Based on this core idea, RMIA introduces a binary relation modeling module to deeply capture the affinity between applicants and approvers, providing a coarse-grained basic information source for approval decisions.

% ==
Specifically, we first use two standard transformers~\cite{vaswani2017attention} to perform self-interaction modeling on the identity information embeddings of applicants and approvers, i.e., $\mathbf{e}^{\mathbf{x}^{id}}$ and $\mathbf{e}^{\mathbf{\overline{x}}^{id}}$, thereby better representing their unique attributes.
\begin{equation}\label{equ:4}
\mathbf{\hat{e}}^{\mathbf{x}^{id}}=\mathrm{Transformer}\left(\mathbf{e}^{\mathbf{x}^{id}}\right),\mathbf{\hat{e}}^{\mathbf{\overline{x}}^{id}}=\mathrm{Transformer}\left(\mathbf{e}^{\mathbf{\overline{x}}^{id}}\right).
\end{equation}
% ==
We then utilize a binary relation extractor to identify intrinsic affinities from the two refined embeddings mentioned above. 
% ==
In practice, we use another standard transformer to complete this process and ultimately obtain the applicant-approver embedding as the basis for subsequent decision making.
\begin{equation}\label{equ:5}
\mathbf{\hat{e}}^{bi}=\mathrm{mean}\left(\mathrm{Transformer}\left(\mathbf{\hat{e}}^{\mathbf{x}^{id}}\oplus\mathbf{\hat{e}}^{\mathbf{\overline{x}}^{id}}\right)\right),
\end{equation}
where $\mathbf{\hat{e}}^{bi}$ is the applicant-approvers embedding that encodes the binary relation, and $\mathrm{mean}(\cdot)$ is the mean pooling operation on the embedding.
% ==
Note that as described in Section~\ref{subsubsec: feature}, we also explicitly introduce relation-enhanced embeddings associated with the affinity between applicants and approvers to assist binary relation modeling in the model optimization process effectively.

\subsubsection{The Ternary Relation Modeling Module}
% ==
After obtaining the basic information for decision-making, the ACFA approval process needs to further construct a ternary relation of applicant-resource-approver with resources as the hub as an effective supplement to the binary relation to provide fine-grained gain information for decision-making.
% ==
On the one hand, even if there is a strong connection between the applicant and the approver, the applicant may request permission for unusual resources. 
% ==
Therefore, it is necessary to integrate the impact of resources on both parties to make a more reasonable decision.
% ==
On the other hand, even if there is a weak connection between the applicant and the approver, the resource the applicant requests permission for may be a common resource across departments or businesses.
% ==
This also requires identifying the impact of resources beyond the binary.
% ==
Based on this core idea, RMIA introduces a ternary relation modeling module to effectively integrate and distinguish the complex coupling relations between applicants, resources, and approvers, providing fine-grained gain information for approval decisions.

% ==
Specifically, we first further obtain the refined embeddings of applicants, resources, and approvers for subsequent ternary relation modeling.
% ==
For resources, we map the resource feature embedding $\mathbf{e^{\mathbf{r}}}$ obtained previously through a fully connected layer to obtain a refined embedding $\mathbf{\hat{e}^{\mathbf{r}}}$.
\begin{equation}\label{equ:6}
\mathbf{\hat{e}}^{\mathbf{r}}=\mathbf{W}_{2}\mathbf{e^{\mathbf{r}}}+\mathbf{b}_{2},
\end{equation}
where $\mathbf{W}_{2}$ and $\mathbf{b}_{2}$ are trainable parameters.
% ==
For applicants, in addition to the refined modeling of identity information embedding (as described in Section~\ref{subsubsec: binary}), their historical behavior embedding $\mathbf{e}^{\mathbf{x}^{hd}}$ is further introduced for refined modeling to characterize the applicants more comprehensively.
% ==
In practice, we use additional standard transformers to self-interact with historical behavior embeddings to identify key behavior patterns and obtain refined embeddings.
\begin{equation}\label{equ:7}
\mathbf{\hat{e}}^{\mathbf{x}^{hd}}=\mathrm{mean}\left(\mathrm{Transformer}\left(\mathbf{e}^{\mathbf{x}^{hd}}\right)\right).
\end{equation}
We then fuse the refined embeddings corresponding to the two different types of information contained in the applicant (i.e., identity information and historical behavior) and use a fully connected layer to obtain the final applicant embedding $\mathbf{\hat{e}}^{\mathbf{x}}$.
\begin{equation}\label{equ:8}
\mathbf{\hat{e}}^{\mathbf{x}}=\mathbf{W}_{3}\left(\mathrm{mean}\left(\mathbf{\hat{e}}^{\mathbf{x}^{id}}\right)\oplus\mathbf{\hat{e}}^{\mathbf{x}^{hd}}\right)+\mathbf{b}_{3},
\end{equation}
where the applicant's two-part refined embedding, $\mathbf{\hat{e}}^{\mathbf{x}^{id}}$ and $\mathbf{\hat{e}}^{\mathbf{x}^{hd}}$, are calculated using Equations~\ref{equ:4} and~\ref{equ:7}, respectively, and $\mathbf{W}_{3}$ and $\mathbf{b}_{3}$ are trainable parameters.
% ==
Similarly, we also refine the approver’s historical behavior embedding, fuse it with the refined identity information embedding, and use a fully connected layer to obtain the final approver embedding $\mathbf{\hat{e}}^{\mathbf{\overline{x}}}$.
\begin{equation}\label{equ:9}
\mathbf{\hat{e}}^{\mathbf{\overline{x}}^{hd}}=\mathrm{mean}\left(\mathrm{Transformer}\left(\mathbf{e}^{\mathbf{\overline{x}}^{hd}}\right)\right),
\end{equation}
\begin{equation}\label{equ:10}
\mathbf{\hat{e}}^{\mathbf{\overline{x}}}==\mathbf{W}_{4}\left(\mathrm{mean}\left(\mathbf{\hat{e}}^{\mathbf{\overline{x}}^{id}}\right)\oplus\mathbf{\hat{e}}^{\mathbf{\overline{x}}^{hd}}\right)+\mathbf{b}_{4},
\end{equation}
where the approver's two-part refined embedding, $\mathbf{\hat{e}}^{\mathbf{\overline{x}}^{id}}$ and $\mathbf{\hat{e}}^{\mathbf{\overline{x}}^{hd}}$, are calculated using Equations~\ref{equ:4} and~\ref{equ:9}, respectively, and $\mathbf{W}_{4}$ and $\mathbf{b}_{4}$ are trainable parameters.

\begin{figure}[htbp]
\centering
\includegraphics[width=1.\linewidth]{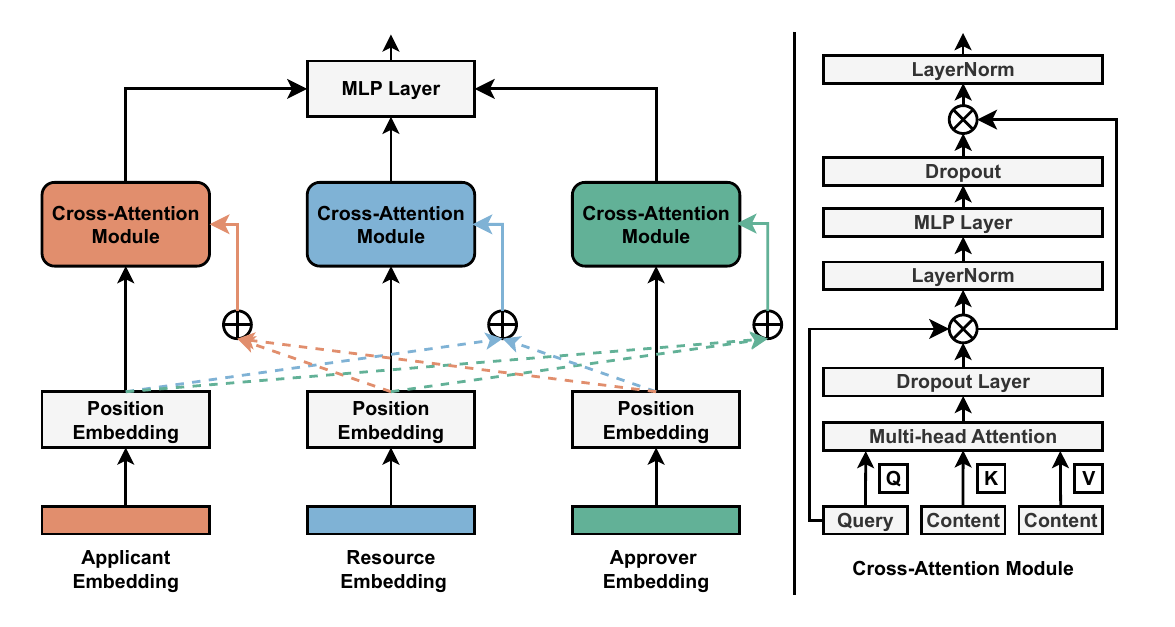}
\caption{Example of a ternary relation extractor.}
\label{fig:3}
\end{figure}

% ==
After obtaining the refined embeddings of applicants, resources, and approvers, i.e., $\mathbf{\hat{e}^{\mathbf{r}}}$, $\mathbf{\hat{e}}^{\mathbf{x}}$, and $\mathbf{\hat{e}}^{\mathbf{\overline{x}}}$, we introduce a ternary relation extractor to effectively mine the complex connections of the three embeddings and obtain the final applicant-resource-approver embedding.
% ==
As shown on the left side of Figure~\ref{fig:3}, we first apply a position embedding to all three of them, then use each of them as the target query in turn, and concatenate the remaining two as auxiliary content.
% ==
We then pass these two pieces of information into a cross-attention module to capture the fusion of the ternary relations from the perspective of each target query.
% ==
The specific calculation of the cross-attention module is shown on the right side of Figure~\ref{fig:3}. 
% ==
In each cross-attention module, one refined embedding will be used as the input of the `Query,' and the concatenation of the other refined embeddings will be used as the input of the rest.
% ==
The overall calculation process can be expressed as follows,
\begin{equation}\label{equ:11}
\mathbf{\overline{e}}^{\mathbf{x}}=\mathrm{Cross\_Attention}\left(\mathbf{\hat{e}}^{\mathbf{x}}+\mathbf{o}^{\mathbf{x}},\left(\mathbf{\hat{e}}^{\mathbf{\overline{x}}}+\mathbf{o}^{\mathbf{\overline{x}}}\right)\oplus\left(\mathbf{\hat{e}}^{\mathbf{r}}+\mathbf{o}^{\mathbf{r}}\right)\right),
\end{equation}
\begin{equation}\label{equ:12}
\mathbf{\overline{e}}^{\mathbf{r}}=\mathrm{Cross\_Attention}\left(\mathbf{\hat{e}}^{\mathbf{r}}+\mathbf{o}^{\mathbf{r}},\left(\mathbf{\hat{e}}^{\mathbf{x}}+\mathbf{o}^{\mathbf{x}}\right)\oplus\left(\mathbf{\hat{e}}^{\mathbf{\overline{x}}}+\mathbf{o}^{\mathbf{\overline{x}}}\right)\right),
\end{equation}
\begin{equation}\label{equ:13}
\mathbf{\overline{e}}^{\mathbf{\overline{x}}}=\mathrm{Cross\_Attention}\left(\mathbf{\hat{e}}^{\mathbf{\overline{x}}}+\mathbf{o}^{\mathbf{\overline{x}}},\left(\mathbf{\hat{e}}^{\mathbf{x}}+\mathbf{o}^{\mathbf{x}}\right)\oplus\left(\mathbf{\hat{e}}^{\mathbf{r}}+\mathbf{o}^{\mathbf{r}}\right)\right),
\end{equation}
where $\mathbf{o}^{\mathbf{x}}$, $\mathbf{o}^{\mathbf{r}}$, and $\mathbf{o}^{\mathbf{\overline{x}}}$ denote the position embeddings corresponding to the applicant, resource, and approver embeddings, respectively.
% ==
$\mathrm{Cross\_Attention(a,b)}$ denotes the modeling process as shown on the right side of Figure~\ref{fig:3}, with the first and second elements serving as the target query and auxiliary content, respectively.

% ==
Next, we concatenate the three embeddings obtained above and use a multi-layer perception (MLP) to get the applicant-resource and approver embeddings corresponding to the ternary relation.
% ==
\begin{equation}\label{equ:14}
\mathbf{\overline{e}}^{te}=\sigma\left(\mathbf{W}_{5}\left(\mathrm{mean}\left(\mathbf{\overline{e}}^{\mathbf{x}}\right)\oplus\mathrm{mean}\left(\mathbf{\overline{e}}^{\mathbf{r}}\right)\oplus\mathrm{mean}\left(\mathbf{\overline{e}}^{\mathbf{\overline{x}}}\right)\right)+\mathbf{b}_{5}\right),
\end{equation}
where $\mathbf{\overline{e}}^{te}$ is the embedding of applicant-resource-approver that encodes the ternary relation; $\mathbf{W}_{5}$ and $\mathbf{b}_{5}$ are trainable parameters; and $\sigma\left(\cdot\right)$ is the activation function.
% ==
Note that as described in Section~\ref{subsubsec: feature}, we additionally explicitly introduce statistical embeddings associated with applicant-resource-approver pairs to effectively assist in ternary relation modeling during the model optimization process.

% =======================================================================================
%=======================================================================================

\subsubsection{The Fusion Decision Module}
% ==
Recall that we have obtained basic text embeddings ($\mathbf{e}^{\mathbf{t}}$) and resource embeddings ($\mathbf{\hat{e}}^{\mathbf{r}}$); applicant-approver embeddings ($\mathbf{\hat{e}}^{bi}$) that reflect the binary relation between applicants and approvers, as well as the auxiliary relation-enhanced embeddings ($\mathbf{e}^{\mathbf{p}}$); applicant-resource-approver embeddings ($\mathbf{\overline{e}}^{te}$) that reflect the complex relation between applicants, resources, and approvers, as well as the auxiliary statistical embeddings ($\mathbf{e}^{\mathbf{s}_n}$ and $\mathbf{e}^{\mathbf{s}_r}$).
% ==
Finally, we concatenate them and use multiple MLP layers to obtain the final approval decision prediction.
\begin{equation}\label{equ:15}
\mathbf{h}=\mathbf{e}^{\mathbf{t}}\oplus\mathbf{\hat{e}}^{\mathbf{r}}\oplus\mathbf{\hat{e}}^{bi}\oplus\mathbf{e}^{\mathbf{p}}\oplus\mathbf{\overline{e}}^{te}\oplus\mathbf{e}^{\mathbf{s}_n}\oplus\mathbf{e}^{\mathbf{s}_r},
\end{equation}
\begin{equation}\label{equ:16}
\hat{y}=\mathrm{softmax}\left(\mathbf{W}_{7}\left(\sigma\left(\mathbf{W}_{6}\mathbf{h}+\mathbf{b}_{6}\right)\right)+\mathbf{b}_{7}\right),
\end{equation}
where $\mathbf{W}_{6}$, $\mathbf{W}_{7}$, $\mathbf{b}_{6}$, and $\mathbf{b}_{7}$ are trainable parameters; $\mathrm{softmax(\cdot)}$ is a softmax operation, since $\hat{y}\in\mathbb{R}^{2 \times 1}$.

%% file: sections/experiments.tex
\section{Experiments}\label{sec:experiments}
% ==
Next, we conduct experiments aimed at answering the following four key questions. 
% ==
The source code will be made public after the paper is accepted.
% ==
\begin{itemize}[leftmargin=*]
    \item \textbf{RQ1}: How does our RMIA perform compared to the baselines?
    \item \textbf{RQ2}: What are the roles of some key steps in our RMIA?
    \item \textbf{RQ3}: What are the characteristics of our RMIA?
    \item \textbf{RQ4}: How does our RMIA perform in real-world ACFA scenarios?
\end{itemize}

\subsection{Experiment Setup}\label{sec:experiments:setup}
\subsubsection{Datasets}
% ==
Since there is a lack of public datasets on intelligent approval of ACFA in the existing literature, we collected and constructed two corresponding experimental datasets from our real-world business, i.e., Business-1 and Business-2.
% ==
The Business-2 dataset is a collection of permission application and approval records from Alibaba's online permission application platform over the past seven months.
% ==
It involves more than 200,000 applicants and approvers, more than 500,000 resources, and provides nearly 100 types of feature information.
% ==
We selected the first six months of data for training and validation, and the last month of data for testing. Because historical approval data is highly imbalanced between positive and negative instances, we retained all negative instances from the first six months. We then randomly sampled an equal number of positive cases. 
% ==
Ultimately, the Business-2 dataset had a training and validation set size of 200,000 and a test set size of 810,000.
% ==
To facilitate research on intelligent approval for ACFAs, we randomly selected over 200 applicants, resources, and approvers from business scenarios.
% ==
We then screened the relevant historical application and approval data to form the Business-1 dataset, which contains more than 100 feature types and 5,000 instances.
% ==
We partitioned this dataset into training, validation, and testing subsets in an 8:1:1 ratio.
% ==
The Business-1 dataset will be publicly available upon acceptance of the paper.

\subsubsection{Metrics}
% ==
Since ACFA's intelligent approval follows the paradigm of a binary classification task, we adopt four widely used binary classification evaluation metrics,  i.e., the area under the ROC curve (AUC), Precision, Recall, and F1-score. 
% ==
We use AUC as the primary metric to guide the optimization process of hyperparameter search in our experiments.

\subsubsection{Baselines}
% ==
To demonstrate the effectiveness of our framework, we compare RMIA with two groups of baselines: machine learning-based classification methods and neural network-based classification methods.
% ==
The baselines of the first group mainly include logistic regression (LR)~\cite{tolles2016logistic}, random forest~\cite{breiman2001random}, LightGBM~\cite{ke2017lightgbm}, XGBoost~\cite{chen2016xgboost}, which are commonly used in the industry, and an ensemble model that combines them with voting mechanisms~\cite{rokach2010ensemble}.
% ==
The second group includes commonly used and recent CTR prediction models in the industry, such as DNN, DeepFM~\cite{guo2017deepfm}, DCN~\cite{wang2021dcn}, EulerNet~\cite{tian2023eulernet}, FinalMLP~\cite{mao2023finalmlp}, and FINAL~\cite{zhu2023final}, which are oriented towards similar binary classification scenarios.
% ==
This makes it easy for us to adapt them to intelligent approval tasks.

\subsubsection{Implementation Details}
% ==
For some general hyperparameters, the embedding dimension and batch size are set to 32 and 128, respectively. 
% ==
For some general settings, we consistently use the AdamW optimizer, batch normalization, and Xavier initialization for all methods, and set the number of iterations to 10.
% ==
We select the optimal learning ratio from $\{$5e-4,1e-4,5e-5,1e-5$\}$ and the $l_2$ regularization from $\{$1e-5, 5e-6, 1e-6, 5e-7, 1e-7$\}$.
% ==
For other benchmark methods, we first consulted their open-source libraries. 
% ==
When open source libraries were lacking, we carefully re-implemented them based on the details provided by the original paper.
% ==
Note that all the methods will search for the best hyperparameter combination within the same search range.
% ==
All experiments are performed on Python 3.10, Torch 2.0.1, and NVIDIA V100.

\subsection{RQ1: Performance Comparison}\label{sec:experiments:rq1}
% == 
% ==
We report the comparison results in Table~\ref{tab: main_result}.
% ==
From the results in Table~\ref{tab: main_result}, we can have the following observations: 
% ==
1) Most methods achieve promising performance results on both datasets, especially when the data size is larger, the overall performance results are better. This shows that with sufficient data and feature information, intelligent approval of ACFA is a feasible and promising direction.
% ==
2) Some machine learning methods (especially tree models) outperformed some neural network methods in both datasets. This may be due to their greater robustness to the high-dimensional sparsity of the ACFA approval data. This advantage is particularly pronounced when the data size is smaller.
% ==
3) The performance differences between most baseline methods are not significant. This indicates that for the ACFA intelligent approval task, traditional binary classification methods that lack targeted modeling will suffer from similar performance bottlenecks.
% ==
4) Our RMIA consistently outperforms all baselines on both datasets and shows significantly better performance gains. This demonstrates the effectiveness of the binary and ternary relation modeling modules introduced in our RMIA for making reasonable decision predictions for ACFA.
% ==
Benefiting from relational modeling and hierarchical decision-making, our RMIA can more effectively identify important influencing factors and incorporate them into the forecasting process.

% =======================================================================================
\begin{table*}[htbp]
     \centering
     \caption{Results on all datasets, where the best results are marked in bold. Note that $^{*}$ indicates a significance level of $p\leq 0.05$ based on a two-sample t-test between our framework and the best baseline.}
     \resizebox{1.0\textwidth}{!}{
         \begin{tabular}{c|cccc|cccc}
         \toprule
            Datasets & \multicolumn{4}{c|}{Business-1} & \multicolumn{4}{c}{Business-2} \\
         \midrule
            Methods & AUC$\uparrow$ & Precision$\uparrow$ & Recall$\uparrow$ & F1-score$\uparrow$ & AUC$\uparrow$ & Precision$\uparrow$ & Recall$\uparrow$ & F1-score$\uparrow$ \\
         \midrule
            LR & 0.7685 & 0.8800 & 0.9600 & 0.9200  & 0.8622 & 0.9900 & 0.6900 & 0.8100\\
            Random Forest & 0.7969 & 0.8700 & 0.8700 & 0.9300 & 0.9232 & 0.9900 & 0.8800 & 0.9300 \\
            LightGBM & 0.8024 & 0.8800 & 0.9800 & 0.9300 & 0.9308 & 0.9900 & 0.8800 & 0.9300 \\
            XGBoost & 0.8052 & 0.8700 & 0.8600 & 0.9300 & 0.9267 & 0.9900 & 0.8700 & 0.9300  \\
            Ensemble & 0.7905 & 0.8800 & 0.9800 & 0.9200 & 0.9209 & 0.9900 & 0.8500 & 0.9200 \\
            DNN & 0.7353 & 0.8700 & 0.8600 & 0.9300 & 0.9171 & 0.9942 & 0.8606 & 0.9226  \\
            DeepFM & 0.6090 & 0.8700 & 0.8900 & 0.8800 & 0.8453 & 0.9918 & 0.8089 & 0.8911  \\
            DCN & 0.7963 & 0.8800 & 0.9700 & 0.9200 & 0.9106 & 0.9934 & 0.8918 & 0.9399  \\
            % CL4CTR & 0.6255 & 0.5500 & 0.6000 & 0.4900 & 0. & 0. & 0. & 0. \\
            EulerNet & 0.7675 & 0.8700 & 0.9800 & 0.9300 & 0.9061 & 0.9900 & 0.9100 & 0.9500 \\
            FinalMLP & 0.7714 & 0.8700 & 0.8700 & 0.9300 & 0.9179 & 0.9900 & 0.9000 & 0.9400 \\
            FINAL & 0.7808 & 0.8700 & 0.8600 & 0.9200 & 0.9085 & 0.9900 & 0.8900 & 0.9400  \\
            RMIA & $\textbf{0.8521}^{*}$ & $\textbf{0.9300}^{*}$ & $\textbf{0.9900}^{*}$ & $\textbf{0.9600}^{*}$ & $\textbf{0.9780}^{*}$ & $\textbf{0.9978}^{*}$ & $\textbf{0.9160}^{*}$ & $\textbf{0.9552}^{*}$  \\
         \bottomrule
         \end{tabular}}%
    \vspace{-10pt}
     \label{tab: main_result}%
\end{table*}%
% =======================================================================================

\subsection{RQ2: Ablation Study}\label{sec:experiments:rq2}
\begin{figure}[htbp]
\centering
\includegraphics[width=.9\linewidth]{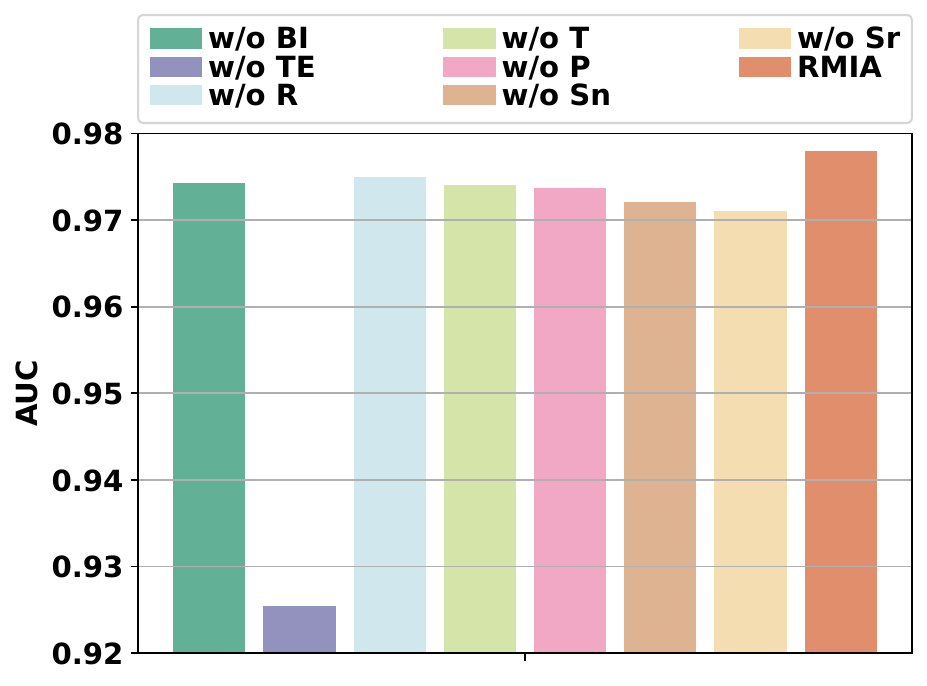}
\caption{AUC results of ablation study on the Business-2 dataset.}
\label{fig:4}
\end{figure}

% == 
To analyze the contribution of some key steps in RMIA, we conduct an ablation study and report the AUC results on the Business-2 dataset in Figure~\ref{fig:4}.
% ==
We evaluated the performance of RMIA when excluding the binary relation modeling module (denoted as ``w/o BI''), excluding the ternary relation modeling module (denoted as ``w/o TE''), excluding resource embeddings (denoted as ``w/o R''), excluding text embeddings (denoted as ``w/o T''), excluding affinity information (denoted as ``w/o P''), excluding historical application number statistics (denoted as ``w/o Sn''), and excluding historical application pass rate statistics (denoted as "w/o Sr").
% ==
Specifically, for these variants, we remove applicant-approver embedding $\mathbf{\hat{e}}^{bi}$, applicant-resource-approver embedding $\mathbf{\overline{e}}^{te}$, resource embedding $\mathbf{\hat{e}}^{\mathbf{r}}$, text embedding $\mathbf{e}^{\mathbf{t}}$, relation-enhanced embedding $\mathbf{e}^{\mathbf{p}}$, and two types of statistical embeddings, $\mathbf{e}^{\mathbf{s}_n}$ and $\mathbf{e}^{\mathbf{s}_r}$, from Equation~\ref{equ:15}, respectively.
% ==
From the results in Figure~\ref{fig:4}, we have the following observations: 
% ==
1) \textbf{``RMIA'' vs. ``w/o BI'', ``w/o P''}.
A variant that removes the binary relation modeling module or affinity information performs worse than RMIA, which indicates the necessity of capturing the affinity relation between applicants and approvers in the intelligent approval of ACFA.
% ==
2) \textbf{``RMIA'' vs. ``w/o TE'', ``w/o Sn'', ``w/o Sr''}.
A variant that removes the ternary relation modeling module or historical application statistics suffers the most significant performance degradation, reflecting the significant impact of the complex coupling relation between applicants, resources, and approvers on ACFA's intelligent approval.
% ==
3) \textbf{``RMIA'' vs. ``w/o R'', ``w/o T''}.
A variant that removes text information also suffers from a performance loss, reflecting that reasonable application text filling can help improve permission application decisions to a certain extent. A variant that removes resource information suffers the least performance loss, possibly because some of its information is already utilized in the ternary relation modeling module.

% =======================================================================================
%=======================================================================================

% % ==================================================================================
% \begin{table}[htbp]
% \centering
% \caption{Ablation Analysis on Ours, where the best results are marked in bold.}
% \resizebox{1.\linewidth}{!}{
%   \begin{tabular}{c|cccc}
%     \specialrule{0.1em}{1pt}{1pt}
%     Metrics & {AUC$\uparrow$} & {Precision$\uparrow$} & {Recall$\uparrow$} & {F1-score$\uparrow$} \\
%     \specialrule{0.05em}{1pt}{3pt}
%     w/o BI & 0.9743 & $\textbf{0.6878}$ & 0.9186 & 0.7540 \\
%     w/o TE & 0.9255 & 0.5773 & 0.8404 & 0.6035 \\
%     w/o R & 0.9750 & 0.7202 & 0.9104 & \textbf{0.7836} \\
%     w/o T & 0.9740 & 0.6314 & 0.9218 & 0.6878 \\
%     w/o P & 0.9737 & 0.6050 & 0.9182 & 0.6485 \\
%     w/o Sn & 0.9721 & 0.6487 & 0.9164 & 0.7098 \\
%     w/o Sr & 0.9710 & 0.6207 & 0.9163 & 0.6726 \\
%     RMIA  & $\textbf{0.9780}$ & 0.9978 & $\textbf{0.9160}$ & 0.9552 \\
%    \specialrule{0.1em}{1pt}{1pt}
%   \end{tabular}}
% \label{tab: ablation_result}
% \end{table}
% % ==================================================================================

\subsection{RQ3: In-depth Analysis of RMIA}\label{subsec:indepth}
% == 
Next, we conduct some in-depth analysis of RMIA's decision-making results, especially to understand which sets of characteristics make permission applications more likely to be approved.
% ==
In the first case, we examined the impact of applicant identity information on approval decisions. The results are shown in Table~\ref{tab:2}, where we use the average approval rate of outsourced employees as an anchor and report the approval rates of other types of applicants. Among them, applications from formal employees have an approval advantage.
%=======================================================================================
\begin{table}[htbp]
  \centering
  \caption{The impact of applicants’ identity information on approval decisions.}
  \resizebox{0.8\linewidth}{!}{
    \begin{tabular}{lccc}
    \toprule
    Categories & Regular Employees & Internship Staff    & Outsourced Employees \\
    \midrule
    Ratio (\%) & +1.93   & +0.85 & 0.00 \\
    \bottomrule
    \end{tabular}}%
  \label{tab:2}%
\end{table}%
%=======================================================================================
% ==
In the second case, we examined the impact of the approver's identity on the approval decision. We divided the approvers into two categories: the applicant's supervisor and ordinary resource approvers, using the average approval rate of ordinary resource approvers as the anchor. The results are shown in Table~\ref{tab:3}. We found that approvers with supervisors who are more likely to be friendly with the applicant have a higher approval rate.
%=======================================================================================
\begin{table}[htbp]
  \centering
  \caption{The impact of approver’s identity information on approval decisions.}
  \resizebox{0.8\linewidth}{!}{
    \begin{tabular}{lcc}
    \toprule
    Categories & Applicant's Supervisor & Ordinary Resource Approvers \\
    \midrule
    Ratio (\%) & +2.86  & 0.00 \\
    \bottomrule
    \end{tabular}}%
  \label{tab:3}%
\end{table}%
%=======================================================================================
% ==
We also analyzed the impact of permission type and found that application approval rates for functional permissions were slightly higher by 0.4\% compared to application approval rates for role permissions. This may be due to the wider range of accessibility granted by role permissions and the higher risk involved.

% ==
We further examined the impact of the affinity between applicants and approvers on approval decisions. We categorized affinity into six levels, with higher levels representing greater affinity. We used the approval rate for the fourth level of affinity as a reference, and the results are reported in Table~\ref{tab:4}. We can see that the impact of affinity is symmetrical. When the affinity between applicants and approvers is higher, the application approval rate is significantly improved. On the other hand, when the affinity between applicants and approvers is lower, we find that the corresponding permission applications are concentrated in the last step of the ACFA, and approvers are more likely to make decisions based on the approvals in the previous steps. This also validates the rationale of introducing relational modeling in RMIA.
%=======================================================================================
\begin{table}[htbp]
  \centering
  \caption{The impact of affinity on approval decisions.}
  \resizebox{0.8\linewidth}{!}{
    \begin{tabular}{lcccccc}
    \toprule
    Categories & C5 & C4 & C3 & C2 & C1 & C0\\
    \midrule
    Ratio (\%) & +5.67 & +4.81 & +3.94 & 0.00 & +2.31 & +4.17\\
    \bottomrule
    \end{tabular}}%
  \label{tab:4}%
\end{table}%
%=======================================================================================
% ==
Finally, we also explored the impact of application text on approval decisions. 
% ==
Finally, we also explored the impact of application text on approval decisions. We can divide application text into three levels based on content: fuzzy descriptions (such as "application for permissions" and "work needs"), general resource descriptions (such as "data analysis" and "viewing data"), and business resource descriptions (such as "development requirements").
% ==
We report the results in Table~\ref{tab:5} using fuzzy descriptions as reference.
% ==
We can find that the application text that is closer to the business needs is more likely to be approved. This provides a good guide for employees to fill in the text when submitting permission applications.
% ==
%=======================================================================================
\begin{table}[htbp]
  \centering
  \caption{The impact of application text on approval decisions.}
  \resizebox{0.9\linewidth}{!}{
    \begin{tabular}{lcccccc}
    \toprule
    Categories & Business Resource Descriptions & General Resource Descriptions & Fuzzy Descriptions\\
    \midrule
    Ratio (\%) & +5.09 & +3.74 & 0.00\\
    \bottomrule
    \end{tabular}}%
  \label{tab:5}%
\end{table}%
%=======================================================================================

\subsection{RQ4: Online Experiments}\label{subsec:online}
\begin{figure}[htbp]
\centering
\includegraphics[width=1.\linewidth]{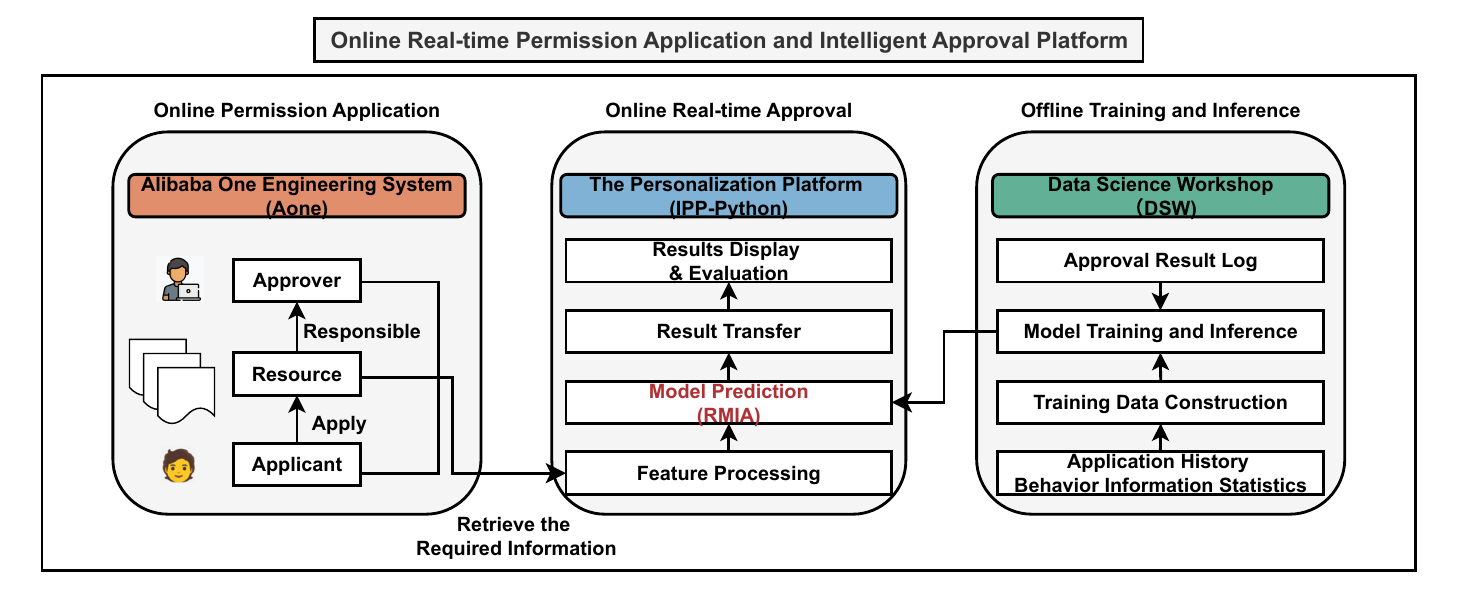}
\caption{Overview of the online ACFA intelligent approval platform.}
\label{fig:5}
\end{figure}

% ==
To verify the effectiveness of our RMIA in real-world applications, we deployed it in Alibaba's online permission application and intelligent approval platform to further evaluate its performance.
% ==
The specific deployment process is shown in Figure~\ref{fig:5}. 
% ==
The platform consists of three main components: online permission application, online real-time approval, and an offline training module.
% ==
When an applicant applies for permission to a specific resource, the platform simultaneously obtains relevant feature information about the applicant, resource, and approver. 
% ==
This information is passed to the decision maker, who uses a pre-trained offline model to generate decision predictions.
% ==
During offline training, the trainer updates the model daily using historical application logs.
% ==
Our RMIA operates offline, leveraging collected application log data to train intelligent approval models for employees across different departments and business scenarios.
% ==
We conducted a month-long online A/B test, randomly selecting 10\% of users as the experimental group and testing the RMIA service. Another 10\% of users were randomly selected as the control group and tested using the baseline model.
% ==
Subsequently, we use several classification metrics that the platform focuses on, AUC, Precision, Recall, and F1-score, as the evaluation metric.
% =======================================================================================
%=======================================================================================

\begin{figure}[htbp]
\centering
\includegraphics[width=0.9\linewidth]{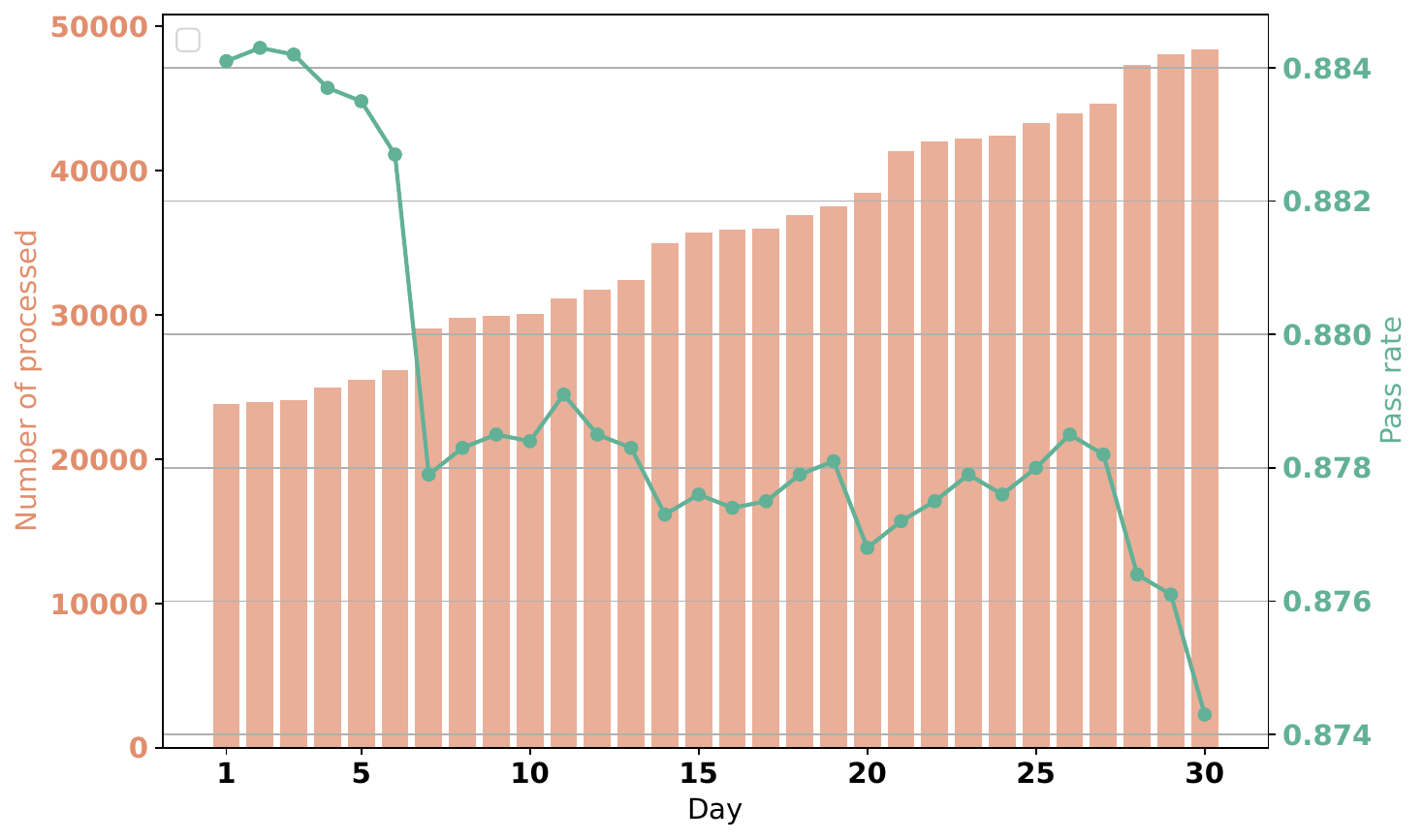}
\caption{Overview of the number of permission requests processed and the model's overall predicted approval rate each day within a month.}
\label{fig:6}
\end{figure}

% ==
As shown in Figure~\ref{fig:6}, we first show the number of applications processed daily by the online permission application and intelligent approval platform and the overall approval rate of RMIA within one month after deploying our RMIA. Note that for confidentiality reasons, we have performed some mapping operations on the specific values. We can observe that our RMIA gradually supports the processing of more permission applications, and the overall predicted trend difference is acceptable (within one percent).
% =======================================================================================
%=======================================================================================
% ==
As shown in Table~\ref{tab:online_result}, our RMIA consistently achieves significant improvements compared to baseline models deployed in online permission application and intelligent approval platforms, and demonstrates the promise of intelligent approval approaches based on relational modeling.
% ==
In addition, after deploying our RMIA, the approval time required for permission applications in different business scenarios was significantly reduced from nearly 10 hours per step to 56 milliseconds, greatly accelerating the management and operational efficiency of various departments.

% =======================================================================================
\begin{table}[htbp]
\centering
\caption{Gains obtained by our RMIA in online deployments.}
\resizebox{0.9\linewidth}{!}{
\begin{tabular}{c|cccc}
\specialrule{0.1em}{3pt}{3pt}
 Metrics & AUC & Precision & Recall & F1-score \\
\specialrule{0.05em}{3pt}{3pt}
Base (LightGBM+BERT+Transformer) & 0.0\% & 0.0\% & 0.0\% & 0.0\%\\
RMIA & +4.90\% & +10.34\% & +8.92\% & 17.77\%\\
\specialrule{0.1em}{3pt}{3pt}
\end{tabular}}
\label{tab:online_result}
\end{table}
% =======================================================================================

%% file: sections/relatedwork.tex
\section{Related Work}\label{sec:related}
% ==
In this section, we briefly review some related work, mainly concerning automated intelligent approval.

% ==
There are diverse approval tasks in various real-world applications.
% ==
With the rapid development of artificial intelligence, introducing artificial intelligence into these approval tasks to automate the approval process, improve approval efficiency, and reduce human errors is a promising research direction~\cite{moscato2021benchmark,hofmann2020robotic}.
% ==
Existing research has developed a series of intelligent approval methods for different application scenarios, such as government approval~\cite{farooq2009intelligent}, financial credit~\cite{chen2016financial}, medical insurance~\cite{johnson2023responsible}, and OA systems~\cite{zhao2024design}, which require the effective integration of different modal data and feature sets contained in these application scenarios.
% ==
However, these OA-oriented approval methods are only targeted at a specific business scenario, such as reimbursement or legal document review, and lack sufficient versatility~\cite{paoki2024artificial}.
% ==
On the other hand, some companies have promoted the intelligent development of their internal approval systems, such as the ByteDance Intelligent Approval System. However, the specific technologies and modeling details used in these systems have not been fully disclosed, and they lack sufficient reproducibility.
% ==
Our RMIA aims to provide effective intelligent approval for common ACFA in OA and will provide reproducible datasets and source code to promote further research on this topic.

%% file: sections/conclusion.tex
\section{Conclusions}\label{sec:conclusions}
% ==
In this paper, we present a novel relational modeling-driven intelligent approval framework, RMIA, for the intelligentization of access control flow approval (ACFA) tasks in office automation (OA) systems.
% ==
Our RMIA decomposes the approval decision into a base part driven by the binary relationship between approver and applicant and a gain part driven by the ternary relationship between approver, resource, and applicant. 
% ==
We then introduce corresponding components to capture these two key relationships and effectively integrate them with other information to make decisions.
% ==
Finally, the effectiveness of our RAMI is verified in two business datasets and an online A/B test.